\documentclass[10pt,twocolumn,letterpaper]{article}

\usepackage[pagenumbers]{cvpr} 

\usepackage{times}
\usepackage{epsfig}
\usepackage{graphicx}
\usepackage{amsmath}
\usepackage{amssymb}
\usepackage{dsfont}
\usepackage{multirow}
\usepackage{enumitem}
\usepackage{booktabs}
\usepackage{colortbl}
\usepackage{relsize}
\usepackage[dvipsnames]{xcolor}

\usepackage[acronym]{glossaries}
\graphicspath{ {./figures/} }
\newacronym{ece}{ECE}{Expected Calibration Error}
\newacronym{coco}{COCO}{Common Objects in Context dataset~\cite{lin2015mscoco}}
\newacronym{tta}{TTA}{Test Time Augmentation}
\newacronym{nms}{NMS}{non-maximum suppression~\cite{rosenfeld1971nms_org}}
\newacronym{tp}{TP}{True Positive}
\newacronym{iou}{IoU}{Intersection over Union}

\newcommand{\shortsec}[1]{\noindent{\textbf{#1}}}
\newcommand{\change}[1]{\fontsize{7pt}{0.1em}\selectfont (\ensuremath#1)}
\newcommand{\variation}[1]{\fontsize{7pt}{0.1em}\selectfont \ensuremath#1}
\newcommand{\map}{\ensuremath{\text{mAP}}\xspace}
\newcommand{\mapfifty}{\ensuremath{\text{mAP}_{50}}\xspace}
\newcommand{\ap}{\ensuremath{\text{AP}}\xspace}
\newcommand{\precision}{\ensuremath{\text{Prec}}\xspace}
\newcommand{\recall}{\ensuremath{\text{Rec}}\xspace}
\newcommand{\iouthresh}{\ensuremath{t_{\text{IoU}}}\xspace}
\newcommand{\detect}[1]{\ensuremath{d_{#1}}\xspace}
\newcommand{\detects}{\ensuremath{\mathcal{D}}\xspace} 
\newcommand{\gtdetects}{\ensuremath{\mathcal{G}}\xspace}
\newcommand{\categ}[1]{\ensuremath{k_{#1}}\xspace}
\newcommand{\bbox}[1]{\ensuremath{b_{#1}}\xspace}
\newcommand{\confidence}[1]{\ensuremath{c_{#1}}\xspace} 
\newcommand{\probtp}[1]{\ensuremath{\mathds{P}_{#1}}\xspace} 
\newcommand{\estprobtp}[1]{\ensuremath{\hat{\probtp{}}_{#1}}\xspace} 
\newcommand{\tp}[1]{\ensuremath{\textit{$T$}_{#1}}\xspace} 
\newcommand{\tpindic}[1]{\ensuremath{\textit{$\tau$}_{#1}}\xspace} 

\newcommand{\calibfunctrue}{\ensuremath{f}\xspace} 
\newcommand{\calibfunc}{\ensuremath{\hat{\calibfunctrue}}\xspace} 
\newcommand{\boxbin}[1]{\ensuremath{B_{#1}}\xspace} 
\newcommand{\confbin}[1]{\ensuremath{C_{#1}}\xspace}

\newcommand{\lossbrier}{\ensuremath{L_{\text{Brier}}}\xspace}
\newcommand{\losslog}{\ensuremath{L_{\log}}\xspace}
\newcommand{\lossmse}{\ensuremath{L_{\widehat{\text{MSE}}}}\xspace}
\newcommand{\lossdiff}{\ensuremath{L_{\text{diff}}}\xspace}
\newcommand{\apest}{\ensuremath{\text{AP}_{\text{est.}}}\xspace}

\newcommand{\E}{\ensuremath{\mathbb{E}}}

\DeclareMathOperator*{\argmax}{arg\,max}

\usepackage[pagebackref=true,breaklinks=true,colorlinks,bookmarks=false]{hyperref}

\usepackage[capitalize]{cleveref}
\crefname{section}{Sec.}{Secs.}
\Crefname{section}{Section}{Sections}
\Crefname{table}{Table}{Tables}
\crefname{table}{Tab.}{Tabs.}

\def\cvprPaperID{****} 
\def\confName{-}
\def\confYear{-}

\begin{document}

\title{The Box Size Confidence Bias Harms Your Object Detector}

\author{
Johannes Gilg\thanks{Correspondence to: Johannes.Gilg@tum.de} \qquad Torben Teepe \qquad Fabian Herzog  \qquad Gerhard Rigoll\vspace{5pt}\\
Technical University of Munich
}

\maketitle
\begin{abstract}
   Countless applications depend on accurate predictions with reliable confidence estimates from modern object detectors.
   It is well known, however, that neural networks including object detectors produce miscalibrated confidence estimates.
   Recent work even suggests that detectors' confidence predictions are biased with respect to object size and position, but it is still unclear how this bias relates to the performance of the affected object detectors.
   We formally prove that the conditional confidence bias is harming the expected performance of object detectors and empirically validate these findings.
   Specifically, we demonstrate how to modify the histogram binning calibration to not only avoid performance impairment but also improve performance through conditional confidence calibration. 
   We further find that the confidence bias is also present in detections generated on the training data of the detector, which we leverage to perform our de-biasing without using additional data.
   Moreover, Test Time Augmentation magnifies this bias, which results in even larger performance gains from our calibration method.
   Finally, we validate our findings on a diverse set of object detection architectures and show improvements of up to 0.6 \map and 0.8 \mapfifty without extra data or training.
   Code available at \footnote{\href{https://github.com/Blueblue4/Object-Detection-Confidence-Bias}{github.com/Blueblue4/Object-Detection-Confidence-Bias}}.
\end{abstract}
\section{Introduction}
Accurate probability estimates are essential for automated decision processes. They are crucial for accurate and reliable performance and for properly assessing risks. This is especially true for object detectors, which are regularly deployed in uniquely critical domains such as automated driving, medical imaging and security applications, where human lives can be at stake.
Despite these high stakes, confidence calibration for object detectors receives comparatively little attention. Most of the attention in the design of object detectors goes toward chasing state-of-the-art results on performance benchmarks, while ignoring problems in the confidence of their predictions. 
Additionally, object detectors have recently been shown to also be vulnerable to conditional confidence biases with respect to their positional regression predictions \cite{kuppers2020multivariate}, but it is still unclear how this bias relates to the performance of the affected object detectors.

\begin{figure}[t] 
       \includegraphics[width={\columnwidth}]{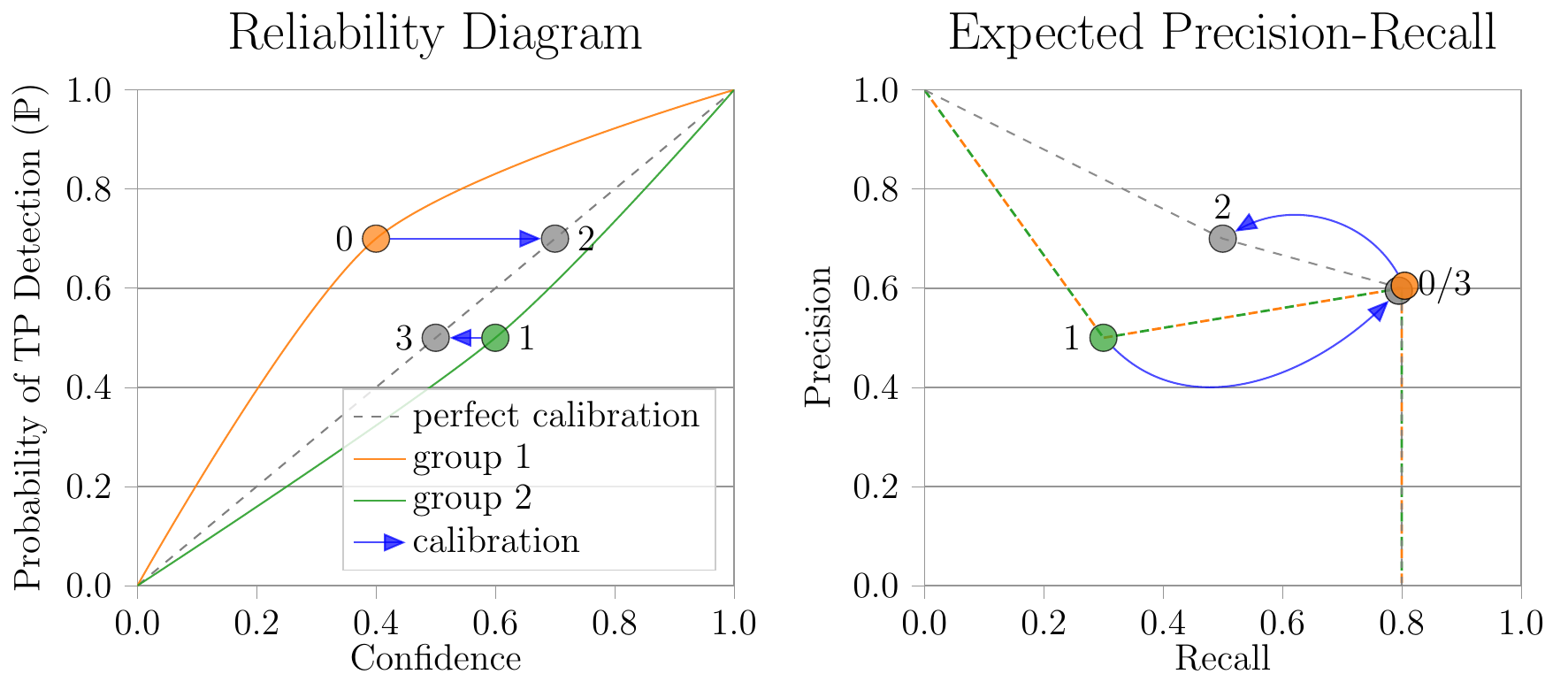}
       \caption{ \textbf{Example illustration of conditional miscalibration hurting object detection performance.} Visualizes, on made up data, how confidence calibration of differently miscalibrated sub-groups, shown (left) in the reliability diagram, increases an object detector's performance, seen (right) in the precision-recall curve.}
    \label{fig:example_ituition}
    \vspace{-5mm}
\end{figure}

In an effort to highlight the importance of confidence calibration, we show that the conditional confidence bias is hurting object detection performance. A simplified illustration of the phenomenon is depicted in \cref{fig:example_ituition}.  

\shortsec{Our Contributions:}
\begin{enumerate}[nosep]

    \item  We formally prove that a conditional bias in object detectors leads to a non-optimal expected Average Precision (\ap) and empirically verify this finding by using a modified histogram binning method.
    \item  We show that the bias is also present in predictions on the training data and the detector can be de-biased thereby increasing its performance without additional data.
    \item  We demonstrate that \gls{tta} can amplify the problems caused by the conditional bias, which causes our remedy to lead to an even larger performance improvement.
    \item  Using our proposed procedure with a heuristic performance metric we are able to improve the performance of most of the tested object detectors on the standard COCO~\cite{lin2015mscoco} evaluation and test-dev benchmark. 
\end{enumerate}
\section{Related Works}
\shortsec{Confidence Calibration of Neural Networks.} 
Confidence calibration is usually applied as a post-processing step for uncertainty estimation. Modern neural networks make highly miscalibrated predictions as shown by Guo \etal~\cite{guo2017on_calib_nn} and hinted at in earlier works~\cite{Nguyen2015easilyfooled, szegedy2014intriguing}. There are many ways to calibrate the confidence of predictive models, such as histogram binning~\cite{zadrozny200histbin}, Bayesian Binning~\cite{naeini2015ece_bayesbin}, isotonic regression \cite{zadrozny2002transforming} and Platt scaling~\cite{platt1999scalingprobabilistic}, with the multi-class modification temperature scaling \cite{guo2017on_calib_nn} and the more general Beta calibration \cite{kull2017betacal}.
Confidence calibration of deep learning object detectors was first addressed by Neumann \etal \cite{neumann2018relaxed} as a learning problem. K\"uppers \etal generalized different calibration methods to conditional calibration of object detectors \cite{kuppers2020multivariate}.
\\
\shortsec{Measuring Calibration Errors.} 
Along with the methods for calibration, measuring to what degree predictions are calibrated is also a long-standing field of study \cite{brier1950verification, Good1952Rational,winkler1968good, degroot1983comparison_reliabilitydiag}. Inspired by the earlier visualization through reliability diagrams \cite{degroot1983comparison_reliabilitydiag}, the nowadays widely used \gls{ece} ~\cite{naeini2015ece_bayesbin} still has many shortcomings exposed and modifications proposed ~\cite{kumar2018trainablecalib, nixon2019measuring_ace, vaicenavicius19evalcal} including an adaption to object detectors~\cite{kuppers2020multivariate}. %
\\
\shortsec{Bias in Deep Learning.}
Bias in deep learning is widely studied, usually in the context of fairness \cite{zemel2013learningfair, pleiss2017fairness, bellamy2018aifairness, huang2019stable, Wang_2020visualbias, sixta2020fairface}, dataset biases \cite{torralba2011unbiased, zhao2017men, alvi2018turning, Kim2021BiaSwap} and learning techniques to mitigate bias during training \cite{Kim2019learning, bahng2020learning, alvi2018turning, zhao2017men, sixta2020fairface}. On the other hand, bias in object detectors is less explored, with the exception of the context bias of object detectors \cite{zhu2019deformable, singh2020contextbias}. Zhao \etal \cite{zhao2017men} have explored label biases in an object detection dataset. K\"uppers \etal~\cite{kuppers2020multivariate} are the first to show conditional bias in the confidence estimates of object detectors with respect to its regressed bounding box outputs. In contrast, we show how the conditional confidence bias with respect to the bounding box is actually detrimental to the performance of object detectors.
\section{Background and Notation} \label{sec:background}
\shortsec{Object Detection.}
An object detector is a predictor that generates a set of detections \detects representing the presence and location of objects in an image. Each of the detector's $N+1=|\detects|$ detections $\detect{i}=(\categ{i}, \bbox{i}, \confidence{i})$, consist of a category \categ{i}, a rectangular bounding box \bbox{i}=$(w,h,x,y)$ and a confidence \confidence{i}. The confidence \confidence{i} represents the  \textit{certainty} of the detector for the presence of an object with the category \categ{i} at the location \bbox{i}.  \\
\shortsec{Evaluating Object Detectors.} 
Object detectors are evaluated against a ground truth set of objects (\gtdetects). The evaluation is performed separately for the detections of every object category. A detection \detect{i} is categorized as a \Gls{tp} if the overlap of its predicted bounding box with a ground truth bounding box is larger than a threshold \iouthresh and if \confidence{i} is highest among all detections that have a large enough overlap with the ground truth bounding box. The overlap is calculated using the Jaccard coefficient, in this context more fittingly termed \Gls{iou}. We define an indicator variable \tpindic{i}, which is $1$ if \detect{i} is a \gls{tp} detection and $0$ otherwise. In the context of object detection, the notion of a True Negative is not well defined as it would correspond to an arbitrary number of ``non-objects" in an image. Therefore, object detectors are evaluated using precision and recall metrics \cite{padilla2020surveymetrics}. 

To compute the precision and recall of an object detector, its detections are sorted according to their confidence from largest to smallest (\confidence{i} $\ge$ \confidence{i+1}, $\forall i \in [1, N-1]$). Then, the precision after $i$ detections $\precision(i)$ is the fraction \gls{tp} predictions \gls{tp}$_i$ out of the $i$ evaluated predictions. By omitting the dependence on \detects, \gtdetects, and \iouthresh for brevity, we can simply write it as
\begin{equation}
	\precision(i) = \frac{\text{TP}_i}{\text{TP}_i+\text{FP}_i} = \frac{\sum_{k=1}^{i} \tpindic{k}}{i}
    \label{eq:prec}
\end{equation}
and, analogously, the recall after $i$ detections is the fraction of \gls{tp} predictions out of the number of available ground truth objects ($|\gtdetects|$):
\begin{equation}
	\recall(i) = \frac{\text{TP}_i}{\text{TP}_i+\text{FN}_i} = \frac{\sum_{k=1}^{i} \tpindic{k}}{|\gtdetects|}.
    \label{eq:rec}
\end{equation}
They can be unified into a single metric - the so-called average precision (\ap)
\begin{equation}
	\ap = \sum_{i=1}^{N} \precision(i)  \cdot \Delta \recall(i),
\label{eq:AP_calculated}
\end{equation}
where $\Delta \recall(i)$ denotes the change of recall from \detect{i-1} to \detect{i}. The \ap is then averaged over a range of $\iouthresh \in [0.50,\,0.55,\,...\,,\, 0.95]$, and over all object categories to get a final mean Average Precision (\map) value, which is a unified performance indicator for a detector. The also used \mapfifty is the class-averaged \ap for $\iouthresh=0.50$. 

Official benchmark implementations of the \map metric apply maximum interpolations of the precision-recall curve and point sampling at specific recall values\cite{lin2015mscoco, everingham2010pascal, padilla2020surveymetrics}. This can produce a slightly more optimistic estimates of the \ap and \map metrics than \cref{eq:AP_calculated}. We also use the official \gls{coco} evaluation script for better comparability on the benchmarks.
\\ 
\shortsec{Confidence Calibration.} 
The idea behind confidence calibration is that the \confidence{i} for each prediction should be equivalent to the empiric object detector's probability for a \Gls{tp} prediction $\probtp{}(\tpindic{i}{=}1|\, \detect{}{=}\detect{i})$. From here on we denote it as \probtp{i} in short. For the confidence calibration, we consider the object detector as a stochastic process. The label of a prediction \detect{i} is now represented by the random variable $\tp{i} \sim \operatorname{Bernoulli}(\probtp{i})$, from which \tpindic{i} with $\iouthresh=0.50$ is drawn as a sample. 
\probtp{i} can also be seen as the precision of the object detector for a set of detections with the same confidence \confidence{i}; we refer to \probtp{i} as probability of a ``successful" or \Gls{tp} detection $\probtp{}(\tpindic{i}{=}1 |\, \detect{}{=}\detect{i})$ to avoid confusion with the metric defined in \cref{eq:prec}. This notation also makes the definition compatible with the confidence calibration of classification neural networks \cite{guo2017on_calib_nn}, as $\probtp{}(\tpindic{i}{=}1)$ is equivalent to the empiric accuracy of a classifier.
Most deep learning-based object detectors are not well calibrated with regard to their confidence estimates \cite{neumann2018relaxed, kuppers2020multivariate}. Therefore, the goal of confidence calibration is to find a mapping \calibfunc that estimates the true confidence calibration curve \calibfunctrue over the input interval $(0,1]$:
 \begin{equation}
    \calibfunctrue(\confidence{i}) = \probtp{}(\tpindic{i}=1 |\confidence{}=\confidence{i}). 
\label{eq:calibrated}
\end{equation}
K\"uppers \etal  discovered that \probtp{i} also depends on the predicted bounding box size and position, not only on \confidence{i}: 
\begin{equation}
    \calibfunctrue(\detect{i}) = \probtp{}(\tpindic{i}=1 |\confidence{}=\confidence{i}, \bbox{}=\bbox{i}) .
\label{eq:cond_calibrated}
\end{equation}
For simplicity, we only focus on the size of the predicted bounding boxes ($h\cdot w$) for the conditional confidence calibration, ignoring the position ($x,y$). The challenge in confidence calibration is that we can only draw from each \tp{i} once. The conditional probability \probtp{} needs to be estimated from the binary outcomes \tpindic{} over all possible confidence values $\confidence{} \in (0,1]$; hence, this is a density estimation problem. 
\\
\shortsec{Histogram Binning.}
One of the most straightforward black-box calibration methods is histogram binning~\cite{zadrozny200histbin}. For histogram binning the predictions are grouped into $M$ confidence intervals \confbin{m} of equal size, so the interval of the $m$-th bin is $\confbin{m} = \left(\frac{m-1}{M},\frac{m}{M}\right]$. The density estimation is performed over the individual intervals separately: The estimated probability of a \gls{tp} detection $\estprobtp{m}$ in interval $m$ is calculated by taking the detections with confidences that lie in the confidence interval and calculating the fraction of detections that are \glspl{tp}. 
The histogram binning calibration of some detection \detect{i} with confidence \confidence{i} is a simple lookup of the calculated average $\estprobtp{\Tilde{m}}$ of the corresponding bin $\confbin{\Tilde{m}} | \confidence{i} \in \confbin{\Tilde{m}}$.
Histogram binning can be extended to a multivariate calibration scheme \cite{kuppers2020multivariate}. For the conditional-dependent binning we first split the detections according to their box size into bins \boxbin{} and then perform the previously described histogram binning for each of the disjoint detection sub-groups. This more general calibration function $\calibfunc_{\confbin{},\boxbin{}}(\detect{})$ produces an estimate for the conditional probability \probtp{}, as described in \cref{eq:cond_calibrated}.
\section{Bias in Confidence of Object Detectors} \label{sec:biasod}
We have the hypothesis that the conditional confidence bias~\cite{kuppers2020multivariate} is hurting object detectors' performance. In \cref{fig:example_ituition} we visualize this idea based on an exaggerated example of two groups of detections  with different calibration curves. Each of the groups only has detections with a single respective confidence value and with this example it is obvious that a detector with a confidence threshold of 0.55 would have a precision of 50\% for the uncalibrated detections (0,1) and a precision of 70\% if the detector was perfectly calibrated (2,3). A related improvement can be observed in the precision recall curve. The area under this curve is closely related to the \ap metric \cite{padilla2020surveymetrics}.
Our simple example and our hypothesis indicate that bias in the confidence estimates of object detectors with respect to bounding box size and position~\cite{kuppers2020multivariate} is hurting the performance of the detectors. However we are interested in a formal proof. 
\subsection{Maximizing Average Precision} \label{subsec:maximizeap}
To prove our assumption that the confidence bias is hurting the performance of object detectors, we take a look at how the \ap relates to \probtp{} and how it can be maximized for a set of detections \detects. An object detector can be seen as a stochastic process (see \cref{sec:background}) so we need to analyze the expected \ap. From \cref{eq:AP_calculated} we get
\begin{equation}
\E_{\tp{}} [\ap] = \E_{\tp{}} \Bigg[ \sum_{i=1}^{N} \precision(i)  \cdot \Delta \recall(i) \Bigg].
\label{eq:exp_ap0}
\end{equation}
Substituting \cref{eq:prec,eq:rec} and our stochastic indicator variable \tp{}, we get:
\begin{equation}
\E_{\tp{}} [\ap] = \E_{\tp{}} \Bigg[ \sum_{i=1}^{N} \bigg( \frac{\sum_{k=1}^{i-1}(\tp{k}) + \tp{i}}{i} \cdot \frac{\tp{i}}{|\gtdetects|} \bigg) \Bigg].
    \label{eq:exp_ap1}
\end{equation}
If we assume independence of \probtp{i} and \probtp{j} for every $i,j$ with $i \neq j$
\begin{equation}
\E_{\tp{}} [\ap] = \frac{1}{|\gtdetects|}  \sum_{i=1}^{N} \bigg(\frac{\sum_{k=1}^{i-1}(\probtp{k}) + 1}{i} \cdot \probtp{i}\bigg).
    \label{eq:exp_ap2}
\end{equation}
With some simple arithmetic we can reformulate this as:
\begin{equation}
\E_{\tp{}} [\ap] =\frac{1}{|\gtdetects|} \sum_{i=1}^{N} \underbrace{\bigg( \frac{\probtp{i}}{i} + \probtp{i} \sum_{k=i+1}^{N}\frac{\probtp{k}}{k} \bigg)}_{h_i(\probtp{i}, \probtp{})}.
    \label{eq:exp_ap3}
\end{equation}
Here, we see that $h_i(l,\probtp{}) > h_{i+1}(l,\probtp{})$ for $i \in \mathbb{N}$ and $l \in (0,1]$. We can therefore maximize the sum in the expected \ap calculation by sorting the predictions according to their \probtp{} from larges to smallest. Since the detections are sorted according to their confidence before evaluating the \ap (see \cref{sec:background}), it is maximized under the following condition:
\begin{equation}
	\probtp{n} < \probtp{m} \; \forall\ n,m\, | \,  \confidence{n} < \confidence{m}.
    \label{eq:ap_maximized}
\end{equation}

It follows that there are only two circumstances under which the calibration can be beneficial for the expected performance of an object detector. The obvious case is when the confidence calibration curve for one class is not monotonic. This is usually not the case, as during training the also monotonic loss function guides the predictions with a empirically higher probability \probtp{i} of being a \gls{tp} to have a higher confidence \confidence{i} through gradient descent, all things being equal.  
The second case is when all things are not equal, as is the case with the conditional confidence bias. When identifiable sub-groups within the predictions of an object detector $\detect{n} \in \detects^A$ and $\detect{m} \in \detects^B$ have different conditional success probability $\probtp{n} \neq \probtp{m}$ for the same confidence $\confidence{n} = \confidence{m}$, this clearly violates \cref{eq:ap_maximized}. For optimal performance the two sub-groups would have to have their confidences \confidence{} transformed so that their combined predictions would have to produce equally precise predictions for equally confident predictions, \ie, have the same monotonic calibration curve. Then their combined detections can satisfy \cref{eq:ap_maximized} which is the required condition to maximize the expected \ap. The interested reader is the referred to \cref{sec:a_max_ap} for a more detailed version of the proof.

\subsection{Confidence Calibration} 
The variation in confidence calibration for different bounding box sizes is reducing the detector's expected performance. Now that we proved this assumption, we want to build on the proof and increase the detector's performance by correcting the variation between calibration curves and see if it increases the performance metrics. The variation is eliminated if we find a mapping for the detection confidences that eliminates the conditional bias, resulting in equal calibration curves \calibfunctrue. This can be reached by mapping the confidences to be equal to their probability of success for each bounding box size. Of course the probability is generally not known, but confidence calibration deals with exactly the problem of finding a function to map confidence scores to their empirical success probability (cf. \cref{sec:background}). 
\\
According to our reasoning, conditional confidence calibration should reduce the box size confidence bias of object detectors. Reducing this bias should increase the \ap of the detector. We try to validate this using the publicly available object detector CenterNet~\cite{zhou2019objects}, with the Hourglass \cite{newell2016HG} backbone, trained on \Gls{coco}. We split the 2017 \Gls{coco} validation set 60:40, calibrate on the first split and evaluate the calibrated detections on the smaller second hold out split. We calibrate class-wise for each of the 80 object categories to account for variations of different categories~\cite{nixon2019measuring_ace} and then split the detections of each class into three equally sized sub-groups \boxbin{} of bounding box sizes. Each sub-group is calibrated using histogram binning with 7 confidence bins \confbin{}.
The calibrated detections perform significantly worse with 35.7 \map than the un-calibrated detections with 40.1 \map (cf. \cref{tab:ablatehistobin}). This result contradicts our initial reasoning and formal proof, what happened?
\\
\shortsec{Modifying Histogram Binning.}
We take a closer look at histogram binning to understand why it drastically reduces the performance of the tested detector. Finding that it violates some of our prior assumptions, we modify the standard histogram binning calibration to actually verify our original hypothesis, that we can use calibration to improve prediction performance. To this end, we infuse histogram binning with the following assumptions. 

\begin{figure}[t] 
        \includegraphics[width={\columnwidth}]{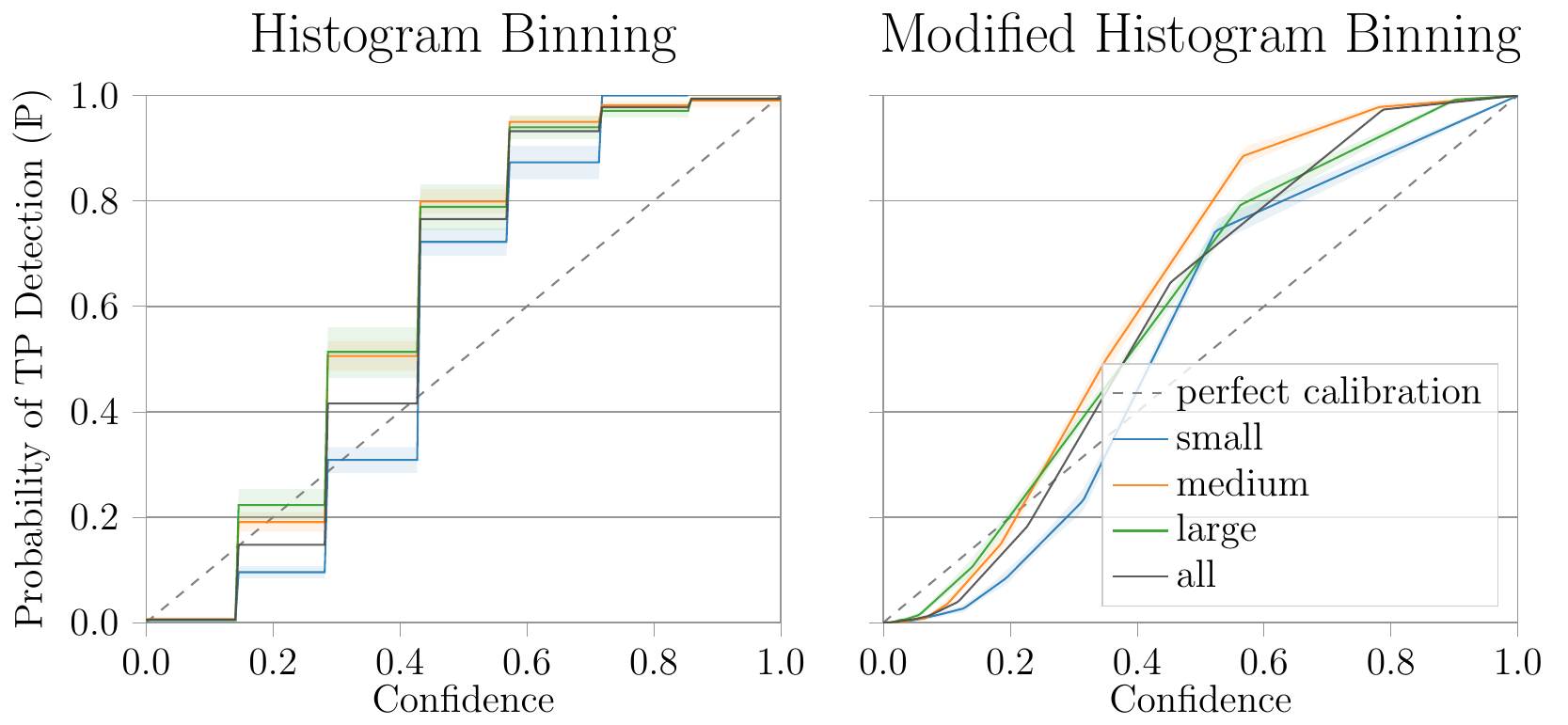}
       \caption{\textbf{Calibration curve of histogram binning and modified version} for category ``person" with 3 size splits and 7 confidence bins (left) and 3 size splits and 14 confidence bins (right), respectively. With the 95\% bootstrap confidence interval for supports. Note that the confidence interval is smaller for the modified version, despite double the number of bins.} 
    \label{fig:histbincurve}
\end{figure}

Our first assumption is that calibration improves our ability to order the predictions according to their probability of being a \gls{tp}. Histogram binning maps confidence ranges to a single estimated precision value, discarding the fine-grained confidence differences (\cf \cref{fig:histbincurve}).
A higher confidence implies a higher probability of being a correct prediction. Since we already split the detections according to their size into sub-groups, we can assume that detectors produce a meaningful confidence ordering within these sub-groups: This is, after all, its training objective. As we want to maintain the ordering within each sub-group we add linear splines between the centers of histogram bins.

This leaves the question of how to extrapolate above the center of the last and below the center of the first bin. We add $\calibfunc(0)=0$ and $\calibfunc(1)=1$ as the outermost interpolation points for the splines to get a mapping of the complete input range $(0, 1]$ to the possible probabilities $(0, 1]$. This encoded prior is also implicit in the beta calibration \cite{kull2017betacal}, as confidences reach their extrema, so should the probability of success: $\lim_{\confidence{} \to 0} \probtp{} = 0$ and $\lim_{\confidence{} \to 1} \probtp{} = 1$. The monotonic mapping is crucial at the boundaries $0$ and $1$ where the confidence values are concentrated, which is desirable property called sharpness~\cite{gneiting2007probab_calib_sharp}. 

The detector's sharpness also leads to problems estimating the probability of a \gls{tp} prediction for each bin. Since the predictions' confidences \confidence{} are non-uniformly distributed, the fixed confidence bins contain a varying amount of detections - sometimes even no detections, which is the case in~\cref{fig:histbincurve} for predictions of small bounding boxes in the confidence interval $0.86 - 1.0$. Due to varying prediction densities some bins with few detections have a high variance. Nixon \etal discovered a similar problem in estimating the calibration error \cite{nixon2019measuring_ace}. We follow their solution and implement a quantile binning scheme. Bin sizes for quantile binning are chosen such that each bin has the same number of predictions, thereby reducing differences in their variance. 

We also set the support for the splines to be at the average confidence of the detections in each bin, to minimize errors from unevenly distributed confidences within each bin. The reduced variance at the supports along with all the modifications can be seen in \cref{fig:histbincurve} (right). We test each modification and the final modified calibration function on the same object detector as before. The results as seen in \cref{tab:ablatehistobin} verify the individual modifications and our original hypothesis, which is that the box size confidence bias reduces the performance of the object detector and our calibration can reduce this bias and increase the performance. 
\begin{table} 
        \centering{\begin{tabular}{lll}
\toprule
Calibration Method      &  \map  &      \mapfifty      \\
\midrule
none                    &     40.10               &   58.82   \\
Histogram Binning       &     35.71\change{-4.39} &   53.25\change{-5.57}   \\
+ Linear Interpolation  &     37.54\change{-2.56} &   55.80\change{-3.02}  \\
+ Added Bounds          &     37.79\change{-2.31} &   56.21\change{-2.61}   \\
+ Adaptive Bins         &     39.84\change{-0.26} &   58.16\change{-0.66}   \\
+ Weighted Supports     &     40.40\change{+0.30} &   59.18\change{+0.36}  \\    
\bottomrule
\end{tabular}

}
        \caption{\textbf{Ablation of histogram binning modifications:} \map and \mapfifty of described modifications for $\boxbin{}=3$ box splits  and $\confbin{}=7$ confidence bins are shown.}
    \label{tab:ablatehistobin}
\end{table}
\subsection{Quantifying Confidence Bias} \label{subsec:measur}
Now that we proved our initial assumption, the question follows of how much improvement can we actually get from calibration. The initial tests were performed by splitting the detections into 3 equally sized bounding box size \boxbin{} sub-groups and using 7 confidence bins \confbin{} to calculate the spline supports. The number of splits are arbitrarily chosen parameters, so there are likely to be better choices for the split sizes since we can change the split sizes for each category. The number of instances per class can vary widely and so can the distributions of bounding box sizes for these instances for different categories. In the \gls{coco} validation dataset split, there are about 250\,k instances for the most frequently occurring category ``person" and fewer than 1000 for the rarest category ``hair dryer". There are bound to be different optima in the bias variance trade-offs for the confidence calibration. There is also the trade-off between the number of box size sub-groups \boxbin{} which enable a more fine grained estimate of the differences in box size-dependent confidence differences and the number of confidence bins \confbin{} that increase the granularity of the calibration curve within a box size split. We need metrics to guide our choice of these parameters for each class. 
\\
\shortsec{Average Precision.} The most obvious objective is to maximize the \ap for each category to get the highest \map on the evaluation split. We explored the connection between the expected \ap and the correct ordering of the detections according to their probability of being a \gls{tp} in \cref{subsec:maximizeap}. As a metric, the \ap can be vulnerable to confidence changes of outlier \gls{tp} detections. 
\\
\shortsec{Proper Scoring Rules.} 
A proper scoring rule is a function $L$ that, in our notation, satisfies the condition $\probtp{i} =\argmax_{\confidence{i}} \E_{\tp{i}} L(\tp{i},\confidence{i})$ \cite{gneiting2007strictly}. Its expected value is maximized for the \Gls{tp} probability \probtp{i} of the Bernoulli random variable \tp{i}. The two most commonly used proper scoring rules are the squared difference, also called the Brier score \cite{brier1950verification} and the logarithmic score \cite{Good1952Rational, gneiting2007strictly}. We derive loss functions by negating the respective scoring functions. The Brier loss is then defined as
\begin{equation}
	\lossbrier = \frac{1}{N} \sum_{i=1}^{N} ( \confidence{i} - \tpindic{i} )^2 ,
    \label{eq:_bier_score}
\end{equation}
while the log loss is defined as
\begin{equation}
	\losslog =  \frac{1}{N} \sum_{i=1}^{N} - \big[ \tpindic{i} \log(\confidence{i}) + ( 1-\tpindic{i} ) \log(1-\confidence{i}) \big] .
    \label{eq:log_loss}
\end{equation}
Both loss functions are minimized for any \confidence{i} only if $\confidence{i} = \probtp{i}$. We do not want to estimate a single \probtp{i}: we want to estimate \probtp{} across the continuous calibration curve \calibfunctrue. Both loss functions are minimized in expectation when \probtp{} is estimated correctly, but the losses favor more confident predictions \cite{winkler1968good}. For each of the \probtp{i} values, deviations and outliers are also penalized by different loss magnitudes. 
\\
\shortsec{Mean Squared Error Estimation.} Ideally, we would like to use an empirically loss function that corresponds to the expected squared error $\E_{\detect{i} \in \detects} [(\calibfunc(\detect{i}) - \probtp{i})^2]$. We can only estimate \probtp{i} from the training data, which brings us right back to the original problem of needing good parameters to estimate the true calibration function \calibfunctrue. The expected error can be split into its bias and variance components: 
\begin{equation} 
    \E_{\detect{} \in \detects} \Big[\big(\calibfunc(\detect{}) - \calibfunctrue(\detect{})\big)^2\Big] = \textit{Bias}_\detects\big(\calibfunc (\detects)\big)^2 + \textit{Var}_\detects\big(\calibfunc (\detects)\big).
    \label{eq:bias_var}
\end{equation}
The variance is easily estimated using K-folds to generate $K$ calibration functions $\calibfunc_{\boxbin{},\confbin{},k}$ and calculating the variance over the population of K calibrated confidences averaged over the entire calibration split. Calculating the actual bias would again require the true calibration function \calibfunctrue, which is unknown. We can, however, estimate how much the bounding box size bias is reduced compared to a non-conditioned calibration scheme $\calibfunc_{1,\confbin{},k}$. We are then able to set the bias to the maximum bias reduction we achieved over the whole explored parameter range $|\boxbin{}| \times |\confbin{}|$. 
\begin{equation} 
\begin{split}
	\widehat{\textit{Bias}}_{\boxbin{},\confbin{}}\big(\calibfunc (\detects)\big) = & \max_{\boxbin{},\confbin{}}\bigg[\frac{1}{N}\sum_{i=1}^{N}\big(\calibfunc_{\boxbin{},\confbin{}}(\detect{i})-\calibfunc_{1,\confbin{}}(\detect{i})\big)\bigg]  \\
	&- \frac{1}{N} \sum_{i=1}^{N}\big(\calibfunc_{\boxbin{},\confbin{}}(\detect{i})-\calibfunc_{1,\confbin{}}(\detect{i})\big).
	\end{split}
    \label{eq:bias_est}
\end{equation}
With this, we can calculate \lossmse as:
\begin{equation} 
    \lossmse = \bigg(\widehat{\textit{Bias}}_{\boxbin{},\confbin{}}\big(\calibfunc (\detects)\big)^2 + \textit{Var}_\detects\big(\calibfunc (\detects)\big)\bigg).
    \label{eq:lossmse}
\end{equation}

\begin{table}
        \centering{\begin{tabular}{lll}
\toprule

Optimized Metric & $\Delta$\map & $\Delta$\mapfifty \\
\midrule
Average Precision ($\mathsmaller \ap$) & $+$0.09\variation{\pm0.10} & $+$0.21\variation{\pm}0.16  \\ 
Brier Loss ($\mathsmaller \lossbrier$) & $-$0.15\variation{\pm0.38} & $-$0.22\variation{\pm0.57}  \\ 
Log Loss ($\mathsmaller \losslog$)     & $+$0.04\variation{\pm0.38} & $+$0.03\variation{\pm0.62} \\  
Est. MSE  ($\mathsmaller \lossmse$)    & $+$0.25\variation{\pm0.18} & $+$0.32\variation{\pm0.28}\\   
\midrule
\textit{oracle}                  & $+$0.67\variation{\pm0.08} & $+$0.97\variation{\pm0.13}  \\ 
\bottomrule
\end{tabular}
%
%
}
        \caption{\textbf{Ablation of optimization metrics of calibration on validation split:} Comparison of performance change after calibrating and optimizing \confbin{} and \boxbin{} with metrics on split of validation data and evaluating on the hold out set, average and max deviation of 10 random splits shown. Optimizing for \lossbrier or \losslog metrics does not improve the \map, or even decreases it.}
    \label{tab:ablateoptimmet}
\end{table}

We test the different metrics on the parameter search space of $\boxbin{} = \{2, 3, 4, 5, 6\}$ and $\confbin{} = \{4, 5, 6, 8, 10, 12, 14\}$ and keep it constant for all following experiments. To our surprise the estimated mean squared error \lossmse performs best among the optimization metrics, even compared to directly evaluating the \ap on the calibration split (\cf \cref{tab:ablateoptimmet}). Still, optimizing for the \lossmse achieves less than half the maximum possible performance gain that is revealed by an \textit{oracle} evaluation on the hold-out split.
\subsection{Bias in Training Predictions}
One objection to our approach could be that we need data beyond the training data of the object detector. For our experiments we used a split of the validation data, but any non-train data would suffice for the calibration. Arguably, the additional data the calibration needs could also be used to improve the detections of the object detector through training, ignoring the benefits of calibrated detections. Motivated by this, we take a look at the object detector's predictions on the training data. 
\begin{figure} 
       \includegraphics[width={\columnwidth}]{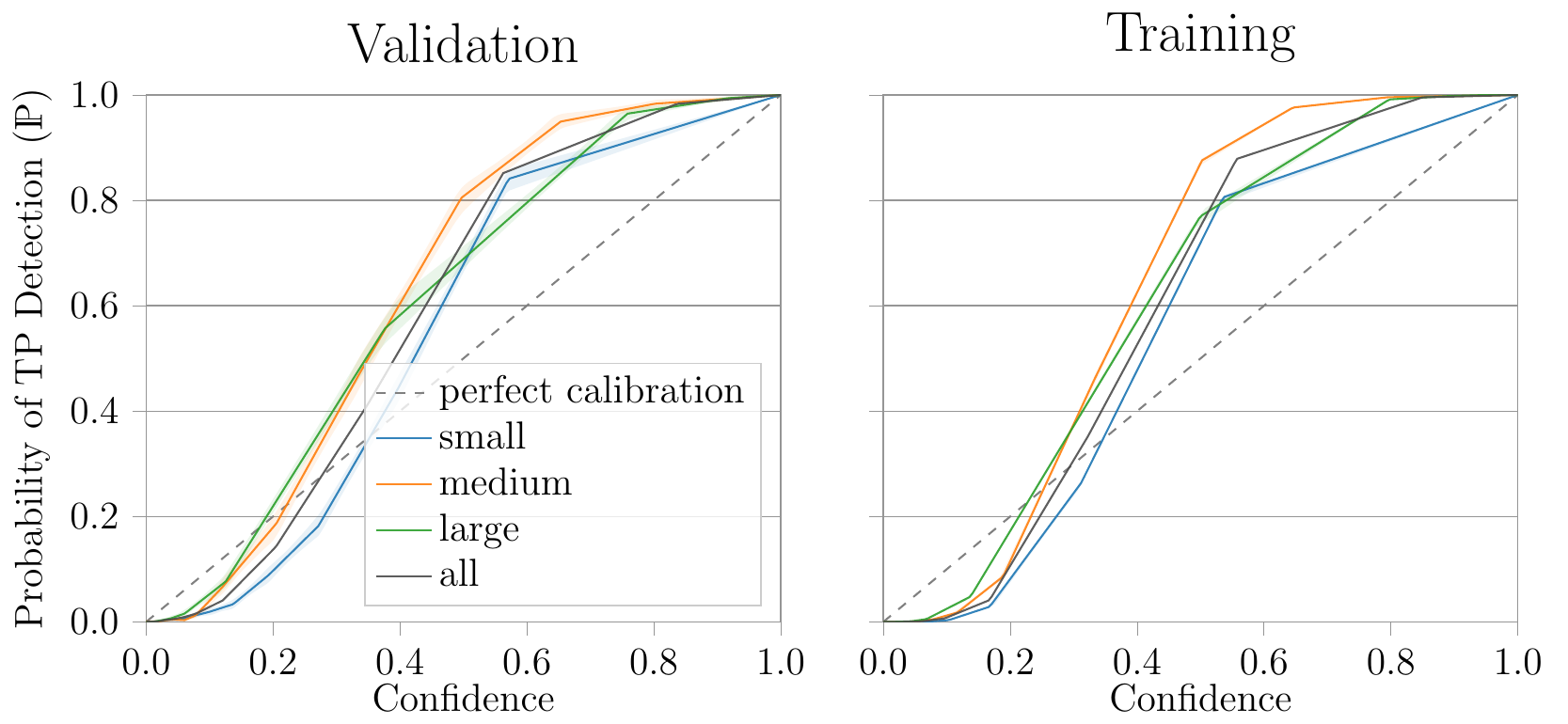}
       \caption{\textbf{Bounding box size bias on train and val data detections} for category "person", with ($\boxbin{}=3$, $\confbin{}=10$) and ($\boxbin{}=3$, $\confbin{}=20$) respectively. Note the 3 size bins here correspond to the official \gls{coco} size splits. Light shaded color represents the 95\% bootstrap confidence interval.}
    \label{fig:sizebiases_train}
\end{figure}
The calibration curves \cref{fig:sizebiases_train} reveal striking similarities between the box-size confidence bias on the training and validation set. They are not identical, but the bias could be similar enough in the predictions training dataset to get a comparable performance gain to the calibration on a validation data split. The calibration curves also show a smaller confidence interval for the training data predictions. The training data with about 118\,k images and 860\,k detections is significantly larger than the validation data with only 5\,k images and about 37\,k detections. The added detections enable a more fine-grained estimation of the reliability diagram, which is indicated by the smaller confidence intervals on the training split (\cf \cref{fig:sizebiases_train}). 

We test the calibration on the detector's predictions on the train set and verify them on the whole validation data. The results are shown in \cref{tab:explore_train} and the performance changes are very similar changes the validation data split (\cf \cref{tab:ablateoptimmet}). 
\begin{table}
        \centering{\begin{tabular}{lll}
\toprule
Optimized Metric & \map & \mapfifty \\
\midrule
none                    & 40.29                 & 59.09 \\
Average Precision ($\mathsmaller \ap$) & 40.25\change{-0.04}   & 59.20\change{+0.11}  \\
Brier Loss ($\mathsmaller \lossbrier$) & 40.36\change{+0.07}   & 59.28\change{+0.19}  \\
Log Loss  ($\mathsmaller \losslog$)    & 40.37\change{+0.08}   & 59.28\change{+0.19} \\
Est. MSE  ($\mathsmaller \lossmse$)    & 40.52\change{+0.23}   & 59.39\change{+0.30} \\
\midrule
\textit{oracle}      & 40.78\change{+0.49}   & 59.78\change{+0.69}  \\
\bottomrule
\end{tabular}}
        \caption{\textbf{Ablation of optimization metrics of calibration on training data.} Calibration and parameter optimization on \gls{coco} training data, evaluated on validation data. Optimization metrics as described in \cref{subsec:measur}. The \lossmse is the best optimization metric and achieves about half the possible performance gain. The \ap performs even worse than on the validation splits and the \lossbrier and \losslog perform slightly better (compare \cref{tab:ablateoptimmet}).}
    \label{tab:explore_train}
\end{table}
\subsection{Test Time Augmentation}

\Gls{tta} is widely used in image recognition tasks to improve prediction performance \cite{szegedy2015going, zhou2019objects}. It is used to produce better and more reliable predictions from a prediction model. The model predictions for an image are generated across different image augmentations, usually geometric transformations, and for object detectors by up- and down-scaling of the images by constant factors. The predictions of the detectors are then combined using some form of \Gls{nms}. 
When the image is down-scaled the model is forced to predict objects with smaller bounding boxes and, according to our reasoning, this should exaggerate the observed confidence bias. As we argued in \cref{subsec:maximizeap}, when differently calibrated sub-groups, the bounding box size groups in this case, are combined the performance is non-optimal and our calibration scheme should improve performance. We can also define the predictions of the detector for one augmentation as a subgroup within all \gls{tta} predictions and calibrate them separately to satisfy \cref{eq:ap_maximized}. Our experiments indeed show, that combining the individually calibrated predictions of each scale augmentation is about three times as effective as only calibrating the combined predictions (\cf \cref{tab:tta}).
\begin{table}
        \centering
        
\begin{tabular}{lcll}

\toprule
Augmentation            & Calib. &\map & \mapfifty \\

\midrule
\multirow{2}{*}{0.50x}  &  -            & 34.45                 &  51.85 \\                 
                        & \checkmark    & 34.70\change{+0.25}   &  52.13\change{+0.28} \\   
\midrule
\multirow{2}{*}{0.75x}  &  -            & 40.82                 &  59.35 \\                 
                        &  \checkmark   & 41.06\change{+0.24}   &  59.62\change{+0.27} \\   
\midrule
\multirow{2}{*}{1.00x}  &  -            & 42.25                 &  61.13 \\                 
                        &  \checkmark   & 42.49\change{+0.24}   &  61.48\change{+0.35} \\   
\midrule
\multirow{2}{*}{1.25x}  &  -            & 40.80                 &  59.98 \\                 
                        &  \checkmark   & 41.08\change{+0.28}   &  60.45\change{+0.47} \\   
\midrule
\multirow{2}{*}{1.50x}  &  -            & 38.53                 &  56.94 \\                 
                        &  \checkmark   & 38.78\change{+0.25}   &  57.39\change{+0.45} \\   
\midrule
\multirow{2}{*}{TTA + NMS}  &  -          & 44.95                 &  64.12 \\                 
                          &  \checkmark & 45.08\change{+0.13}   &  64.30\change{+0.18} \\   
\midrule
cal.(TTA) + NMS             & (\checkmark) & 45.43\change{+0.48}  &  64.64\change{+0.52} \\   
\bottomrule

\end{tabular}
        \caption{\textbf{Effect of individual calibration on \gls{tta},} calibration on \gls{coco} train, evaluated on validation data. (\checkmark): Calibrated  predictions individually for each augmentation.}
    \label{tab:tta}
\end{table}

\section{Evaluation and Discussion} \label{sec:evaluation}

\begin{figure}[t] 
      \includegraphics[width={\columnwidth}]{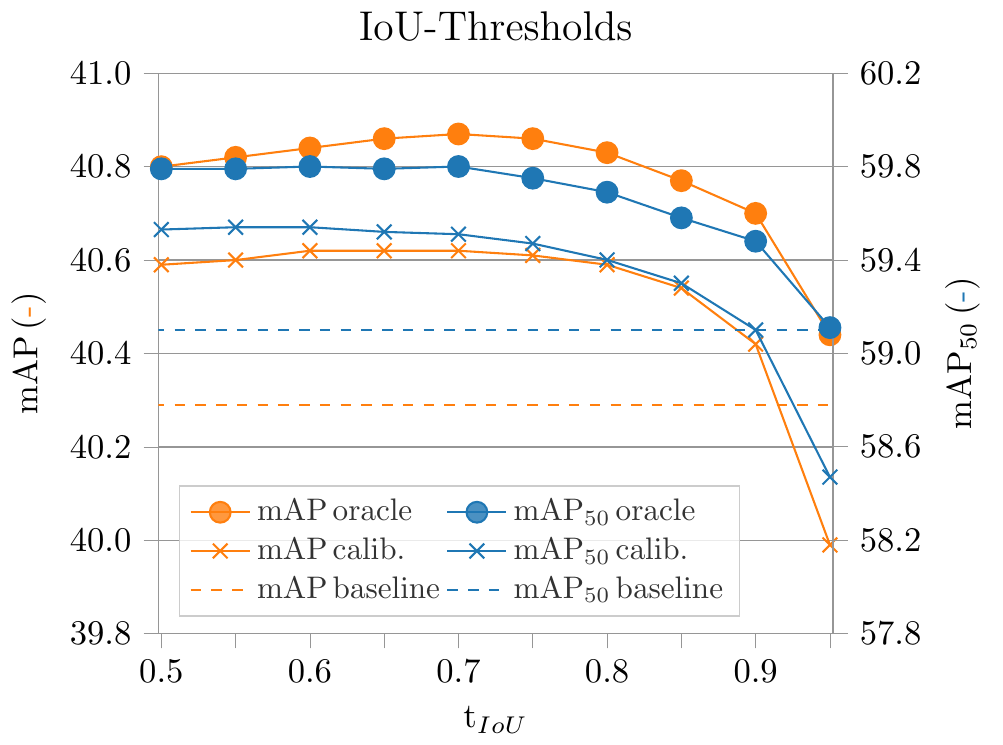}
      \caption{\textbf{Ablation of \gls{iou} thresholds,} performance of conditionally calibrated CenterNet for different values of \iouthresh that determine the required bounding box overlap for \gls{tp} detections.} 
    \label{fig:ablate_iou}
\end{figure}
Finally, we tested our calibration method on a wide range of deep learning object detectors. We kept the parameter search space as before, calibrate on the \gls{coco} train split, and evaluated on the validation and test-dev splits. The results are shown in \cref{tab:modelsperform}. The performance changes vary across the different architectures, ranging from a slight decrease of 0.1 \map and \mapfifty to a large gain of 0.6 \map and 0.8 \mapfifty. There are a variety of influences and limitations on the performance gains.\\
\begin{table*}[ht]
     \centering{\newcolumntype{s}{>{\columncolor[HTML]{EDEDED}} l}
\begin{tabular}{llllsls}
\toprule
\multirow{2}{*}{Detector}   & \multirow{2}{*}{Backbone}     & \multirow{2}{*}{Calibration} & \multicolumn{2}{c}{Validation} & \multicolumn{2}{c}{Test-Dev}   \\\cmidrule(l){4-5} \cmidrule(l){6-7}
                            &                               &      &   \map   &   \mapfifty  &   \map   &   \mapfifty     \\ 
\midrule\midrule
\multirow{3}{*}{YOLOv5X \cite{Glenn2020YOLOv5}}   & \multirow{3}{*}{CSPDarkNet-53 \cite{redmon2018yolov3,wang2020cspnet}} & -      & 50.38 & 68.76 & 50.3 & 68.4 \\ 
                        &                           & train     & 50.42\change{+0.04} & 68.81\change{+0.05} & 50.3\change{+0.0} & 68.4\change{+0.0} \\
                        &                           & \textit{oracle}    & 50.63\change{+0.25} & 69.04\change{+0.28} & - & - \\ 
\midrule
\multirow{3}{*}{RetinaNet \cite{lin2017retinanet}\textdagger} & \multirow{3}{*}{ResNeXt-101 \cite{xie2017resnext}}                                                                      & -  & 40.82               & 60.48 & 41.2 & 61.1 \\
                        &                       & train     & 40.79\change{-0.03} & 60.53\change{+0.05} & 41.2\change{+0.0} & 61.2\change{+0.1} \\
                        &                       & \textit{oracle}    & 41.09\change{+0.27} & 60.76\change{+0.28} & - & - \\
\midrule
\multirow{3}{*}{CenterNet \cite{zhou2019objects}}   & \multirow{3}{*}{Hourglass-104 \cite{newell2016HG}}   & - & 40.29 & 59.10 & 40.2 & 59.1 \\ 
                        &                       & train     & 40.59\change{+0.30} & 59.53\change{+0.43} & 40.5\change{+0.2} & 59.5\change{+0.4} \\
                        &                       & \textit{oracle}    & 40.80\change{+0.51} & 59.85\change{+0.75} & - & - \\ 
\midrule
\multirow{3}{*}{DETR \cite{carion2020detr}\textdagger}   & \multirow{3}{*}{ResNet-50 \cite{he2016deepres}}   & -      & 40.11               & 60.62 & 39.9 & 60.7 \\
                        &                                                   & train     & 40.39\change{+0.28} & 60.68\change{+0.06} & 40.3\change{+0.4} & 60.8\change{+0.1} \\
                        &                                                   & \textit{oracle}    & 40.72\change{+0.61} & 60.86\change{+0.24} & - & - \\
\midrule
\multirow{3}{*}{CenterNet \cite{zhou2019objects}\textdagger}   & \multirow{3}{*}{ResNet-18 \cite{he2016deepres}}   & - & 29.55 & 46.14 & 29.7  & 46.7 \\
                        &                       & train     & 30.06\change{+0.51} &  46.87\change{+0.73} & 30.3\change{+0.6} &  47.5\change{+0.8} \\
                        &                       & \textit{oracle}    & 30.34\change{+0.79} &  47.20\change{+1.06} & - & - \\
\midrule
\multirow{3}{*}{YOLOv3-320 \cite{redmon2018yolov3}\textdagger} & \multirow{3}{*}{DarkNet-53 \cite{redmon2018yolov3}}& -   & 27.91 & 49.10 & 27.8 & 49.0 \\
                        &                                                                                     & train  & 28.22\change{+0.31} & 49.65\change{+0.55} & 28.1\change{+0.3} & 49.5\change{+0.5} \\
                        &                                                                                     & \textit{oracle} & 28.54\change{+0.63} & 49.95\change{+0.85} & - & - \\
\midrule
\multirow{3}{*}{SSD300 \cite{Liu2016ssd}\textdaggerdbl} & \multirow{3}{*}{ResNet-50 \cite{he2016deepres}} & -      & 25.03               & 42.33               & 24.9              & 42.5 \\
                        &                       & train     & 25.17\change{+0.14} & 42.58\change{+0.25} & 24.9\change{+0.0} & 42.7\change{+0.2} \\
                        &                       & \textit{oracle}    & 25.34\change{+0.31} & 42.83\change{+0.50} & - & - \\ 
\midrule\midrule
\multirow{3}{*}{CenterNet2 \cite{zhou2021cnet2}}   & \multirow{3}{*}{ResNet-50 \cite{he2016deepres}}  & -      & 42.86               & 59.52               & 43.1 & 59.9 \\
                        &                               & train     & 42.84\change{-0.02} & 59.50\change{-0.02} & 43.1\change{+0.0} & 59.9\change{+0.0} \\
                        &                               & \textit{oracle}    & 43.08\change{+0.22} & 59.74\change{+0.22} & - & - \\

\midrule
\multirow{3}{*}{Faster-RCNN \cite{Ren2017fasterrcnn}\textdagger}  & \multirow{3}{*}{ResNeXt-101 \cite{xie2017resnext}}  & -         & 41.60                 & 61.93 & 42.0 & 62.8 \\
                        & & train   & 41.52\change{-0.08}   & 61.90\change{-0.03} & 41.9\change{-0.1} & 62.7\change{-0.1}\\
                        & & \textit{oracle}  & 41.80\change{+0.20}   & 62.06\change{+0.13} & - & - \\
\bottomrule
\end{tabular}
}
    \caption{\textbf{Calibration method on different models.}  Calibration and parameter optimization on \gls{coco} train, evaluated on validation data and test-dev benchmark. Models are sorted by performance and separated into one- and two-stage architectures. Official implementation are evaluated unless noted as: \textdagger: \cite{Chen2019mmdetection}, \textdaggerdbl: \cite{pasyke2019pytorch}}
    \label{tab:modelsperform}
\end{table*}
\shortsec{Two Stage vs. One Stage.} 
Our approach does not perform well on both two-stage detectors, even with the \textit{oracle} parameter choice there are little performance changes (\cf \cref{tab:modelsperform}). When the region proposal is separated from the classification and confidence prediction, as in two-stage detectors, it makes sense that it is less likely that a conditional bias on the confidence predictions with respect to the bounding box values is introduced. \\
\shortsec{Model Size.}
The effectiveness of our approach seems to be negatively correlated with the object detectors model size and performance. To verify this trend we test our calibration method on the popular EfficientDet \cite{tan2020efficientdet}, which is available in 8 different size and performance implementation. The trend also holds within this architecture type across the model size scales, with minor outliers (\cf \cref{tab:modelsize} for exact values). The larger models appear to learn to reduce the conditional bias to some extent. 
\\
\shortsec{\gls{iou} threshold.} The calibrations are all performed with \iouthresh = 0.5; we analyze the influence of this choice in \cref{fig:ablate_iou}. Unsurprisingly, the \mapfifty metric is maximized for \iouthresh of around 0.5, since this is also the threshold used for identifying \gls{tp} detection. Higher performance in \map can be achieved with $\iouthresh \approx 0.7$ since this is roughly the median of the threshold range for the evaluation of the \map, as described in \cref{sec:background}.
\\
\shortsec{Parameters.}  
We chose a very limited parameter search space for \confbin{} and \boxbin{} and kept it constant throughout all our experiments to keep the results comparable. The chosen space is not guaranteed to contain the optimal trade-off for all, or even any of the object categories and detector architectures. In \cref{sec:a_search_space} we take a look at different search spaces, with even better results for CenterNet. On the other hand, larger number of bins, with not enough detections to get an accurate estimate of \probtp{}, will also give worse results. The conservatively chosen search space and proposed evaluation heuristic \lossmse offer only a first look at the connection of conditional biases and performance.
\\
\shortsec{Calibration.} Detectors that are conditionally calibrated on the training data should not be considered well calibrated since the object detectors performance on the training data usually overestimates its performance on unseen data. For a calibration beyond the reduced confidence bias, it has to be performed on a hold-out set.\\
\shortsec{Broader Impact.} Object detectors are part of many real-world systems, most of which have a positive impact, but there are also many systems with negative societal impacts. These harms can be caused by unintended biases or intentionally, as in autonomous weapons systems. Our proposed de-biasing can be used to increase the performance of object detectors, so it can clearly be used to increase the harm of intentionally designed harmful systems. We believe, however, that our research will have a net-positive impact, as it can improve applications with a positive impact and also reduce unintended harms caused by biased and uncalibrated predictions. 
%
\section{Conclusion} 
We formally proved that the conditional confidence bias is non-optimal for object detectors' performance. We also demonstrated how a slightly modified version of the popular histogram binning can be leveraged to compensate this bias and improve the performance of the object detectors even when calibrated on the training data. While this performance boost varies for different architectures, it is achieved without extra training time, data, or model parameters, and it highlights the fact that the confidence bias is severe enough to have a real performance impact. The proposed de-biasing is especially effective in \gls{tta} and on smaller models that are widely deployed on mobile and edge devices. Our formal and empiric results linking conditional bias to object detectors performance demonstrate the crucial need for researchers and practitioners to pay closer attention to conditional biases and confidence calibration.

{\small
\bibliographystyle{ieee_fullname}
\bibliography{paperbib}

\begin{thebibliography}{10}\itemsep=-1pt

\bibitem{alvi2018turning}
Mohsan Alvi, Andrew Zisserman, and Christoffer Nell{\aa}ker.
\newblock Turning a blind eye: Explicit removal of biases and variation from
  deep neural network embeddings.
\newblock In {\em ECCVW}, pages 556--572, 2018.

\bibitem{bahng2020learning}
Hyojin Bahng, Sanghyuk Chun, Sangdoo Yun, Jaegul Choo, and Seong~Joon Oh.
\newblock Learning de-biased representations with biased representations.
\newblock In {\em ICML}, pages 528--539, 2020.

\bibitem{bellamy2018aifairness}
Rachel~KE Bellamy, Kuntal Dey, Michael Hind, Samuel~C Hoffman, Stephanie Houde,
  Kalapriya Kannan, Pranay Lohia, Jacquelyn Martino, Sameep Mehta, Aleksandra
  Mojsilovic, et~al.
\newblock Ai fairness 360: An extensible toolkit for detecting, understanding,
  and mitigating unwanted algorithmic bias.
\newblock {\em arXiv preprint arXiv:1810.01943}, 2018.

\bibitem{brier1950verification}
Glenn~W Brier et~al.
\newblock Verification of forecasts expressed in terms of probability.
\newblock {\em Monthly weather review}, 78(1):1--3, 1950.

\bibitem{carion2020detr}
Nicolas Carion, Francisco Massa, Gabriel Synnaeve, Nicolas Usunier, Alexander
  Kirillov, and Sergey Zagoruyko.
\newblock End-to-end object detection with transformers.
\newblock In {\em ECCV}, pages 213--229, 2020.

\bibitem{Chen2019mmdetection}
Kai et~al. Chen.
\newblock {MMDetection}: Open mmlab detection toolbox and benchmark.
\newblock {\em arXiv preprint arXiv:1906.07155}, 2019.
\newblock {(Apache-2.0)}: \url{https://github.com/open-mmlab/mmdetection}.

\bibitem{degroot1983comparison_reliabilitydiag}
Morris~H DeGroot and Stephen~E Fienberg.
\newblock The comparison and evaluation of forecasters.
\newblock {\em Journal of the Royal Statistical Society: Series D (The
  Statistician)}, 32(1-2):12--22, 1983.

\bibitem{everingham2010pascal}
Mark Everingham, Luc Van~Gool, Christopher~KI Williams, John Winn, and Andrew
  Zisserman.
\newblock The pascal visual object classes (voc) challenge.
\newblock {\em IJCV}, 88(2):303--338, 2010.

\bibitem{gneiting2007probab_calib_sharp}
Tilmann Gneiting, Fadoua Balabdaoui, and Adrian~E Raftery.
\newblock Probabilistic forecasts, calibration and sharpness.
\newblock {\em Journal of the Royal Statistical Society: Series B (Statistical
  Methodology)}, 69(2):243--268, 2007.

\bibitem{gneiting2007strictly}
Tilmann Gneiting and Adrian~E Raftery.
\newblock Strictly proper scoring rules, prediction, and estimation.
\newblock {\em Journal of the American statistical Association},
  102(477):359--378, 2007.

\bibitem{Good1952Rational}
Irving~J Good.
\newblock Rational decisions.
\newblock {\em Journal of the Royal Statistical Society. Series B
  (Methodological)}, 14(1):107--114, 1952.

\bibitem{guo2017on_calib_nn}
Chuan Guo, Geoff Pleiss, Yu Sun, and Kilian~Q Weinberger.
\newblock On calibration of modern neural networks.
\newblock In {\em ICML}, pages 1321--1330, 2017.

\bibitem{he2016deepres}
Kaiming He, Xiangyu Zhang, Shaoqing Ren, and Jian Sun.
\newblock Deep residual learning for image recognition.
\newblock In {\em CVPR}, pages 770--778, 2016.

\bibitem{huang2019stable}
Lingxiao Huang and Nisheeth Vishnoi.
\newblock Stable and fair classification.
\newblock In {\em ICML}, pages 2879--2890, 2019.

\bibitem{Glenn2020YOLOv5}
Glenn Jocher.
\newblock {YOLOv5}, 2020.
\newblock {(GNU GPL)}: \url{https://github.com/ultralytics/yolov5}.

\bibitem{Kim2019learning}
Byungju Kim, Hyunwoo Kim, Kyungsu Kim, Sungjin Kim, and Junmo Kim.
\newblock Learning not to learn: Training deep neural networks with biased
  data.
\newblock In {\em CVPR}, pages 9012--9020, 2019.

\bibitem{Kim2021BiaSwap}
Eungyeup Kim, Jihyeon Lee, and Jaegul Choo.
\newblock Biaswap: Removing dataset bias with bias-tailored swapping
  augmentation.
\newblock In {\em ICCV}, pages 14992--15001, 2021.

\bibitem{kull2017betacal}
Meelis Kull, Telmo Silva~Filho, and Peter Flach.
\newblock Beta calibration: a well-founded and easily implemented improvement
  on logistic calibration for binary classifiers.
\newblock In {\em Artificial Intelligence and Statistics}, pages 623--631,
  2017.

\bibitem{kumar2018trainablecalib}
Aviral Kumar, Sunita Sarawagi, and Ujjwal Jain.
\newblock Trainable calibration measures for neural networks from kernel mean
  embeddings.
\newblock In {\em ICML}, volume~80, pages 2805--2814, 2018.

\bibitem{kuppers2020multivariate}
Fabian K{\"u}ppers, Jan Kronenberger, Amirhossein Shantia, and Anselm
  Haselhoff.
\newblock Multivariate confidence calibration for object detection.
\newblock In {\em CVPRW}, pages 326--327, 2020.

\bibitem{lin2017retinanet}
Tsung-Yi Lin, Priya Goyal, Ross Girshick, Kaiming He, and Piotr Doll{\'a}r.
\newblock Focal loss for dense object detection.
\newblock In {\em ICCV}, pages 2980--2988, 2017.

\bibitem{lin2015mscoco}
Tsung-Yi Lin, Michael Maire, Serge Belongie, James Hays, Pietro Perona, Deva
  Ramanan, Piotr Doll{\'a}r, and C~Lawrence Zitnick.
\newblock Microsoft coco: Common objects in context.
\newblock In {\em European conference on computer vision}, pages 740--755,
  2014.
\newblock {(CC-BY 4.0)}: \url{https://cocodataset.org}.

\bibitem{Liu2016ssd}
Wei Liu, Dragomir Anguelov, Dumitru Erhan, Christian Szegedy, Scott Reed,
  Cheng-Yang Fu, and Alexander~C. Berg.
\newblock Ssd: Single shot multibox detector.
\newblock In {\em ECCV}, pages 21--37, 2016.

\bibitem{naeini2015ece_bayesbin}
Mahdi~Pakdaman Naeini, Gregory Cooper, and Milos Hauskrecht.
\newblock Obtaining well calibrated probabilities using bayesian binning.
\newblock In {\em AAAI}, 2015.

\bibitem{neumann2018relaxed}
Lukas Neumann, Andrew Zisserman, and Andrea Vedaldi.
\newblock Relaxed softmax: Efficient confidence auto-calibration for safe
  pedestrian detection.
\newblock In {\em NeurIPSW}, 2018.

\bibitem{newell2016HG}
Alejandro Newell, Kaiyu Yang, and Jia Deng.
\newblock Stacked hourglass networks for human pose estimation.
\newblock In {\em ECCV}, pages 483--499, 2016.

\bibitem{Nguyen2015easilyfooled}
Anh Nguyen, Jason Yosinski, and Jeff Clune.
\newblock Deep neural networks are easily fooled: High confidence predictions
  for unrecognizable images.
\newblock In {\em CVPR}, 2015.

\bibitem{nixon2019measuring_ace}
Jeremy Nixon, Michael~W Dusenberry, Linchuan Zhang, Ghassen Jerfel, and Dustin
  Tran.
\newblock Measuring calibration in deep learning.
\newblock In {\em CVPRW}, volume~2, 2019.

\bibitem{padilla2020surveymetrics}
Rafael Padilla, Sergio~L. Netto, and Eduardo A.~B. da Silva.
\newblock A survey on performance metrics for object-detection algorithms.
\newblock In {\em International Conference on Systems, Signals and Image
  Processing (IWSSIP)}, pages 237--242, 2020.

\bibitem{pasyke2019pytorch}
Adam et~al. Paszke.
\newblock Pytorch: An imperative style, high-performance deep learning library.
\newblock In {\em NeurIPS}, pages 8024--8035, 2019.
\newblock {(BSD-3)}: \url{https://pytorch.org/}.

\bibitem{platt1999scalingprobabilistic}
John Platt.
\newblock Probabilistic outputs for support vector machines and comparisons to
  regularized likelihood methods.
\newblock {\em Advances in large margin classifiers}, 10(3):61--74, 1999.

\bibitem{pleiss2017fairness}
Geoff Pleiss, Manish Raghavan, JonKleinberg FelixWu, and KilianQ Weinberger.
\newblock On fairness and calibration.
\newblock In {\em NeurIPS}, pages 5680--5689, 2017.

\bibitem{redmon2018yolov3}
Joseph Redmon and Ali Farhadi.
\newblock Yolov3: An incremental improvement.
\newblock {\em arXiv preprint arXiv:1804.02767}, 2018.

\bibitem{Ren2017fasterrcnn}
Shaoqing Ren, Kaiming He, Ross Girshick, and Jian Sun.
\newblock Faster r-cnn: Towards real-time object detection with region proposal
  networks.
\newblock {\em PAMI}, 28:91--99, Jun 2017.

\bibitem{rosenfeld1971nms_org}
Azriel Rosenfeld and Mark Thurston.
\newblock Edge and curve detection for visual scene analysis.
\newblock {\em IEEE Transactions on Computers}, 100(5):562--569, 1971.

\bibitem{singh2020contextbias}
Krishna~Kumar Singh, Dhruv Mahajan, Kristen Grauman, Yong~Jae Lee, Matt
  Feiszli, and Deepti Ghadiyaram.
\newblock Don't judge an object by its context: Learning to overcome contextual
  bias.
\newblock In {\em CVPR}, pages 11070--11078, 2020.

\bibitem{sixta2020fairface}
Tom{\'a}{\v{s}} Sixta, Julio CS~Jacques Junior, Pau Buch-Cardona, Eduard
  Vazquez, and Sergio Escalera.
\newblock Fairface challenge at eccv 2020: Analyzing bias in face recognition.
\newblock In {\em ECCV}, pages 463--481, 2020.

\bibitem{szegedy2015going}
Christian Szegedy, Wei Liu, Yangqing Jia, Pierre Sermanet, Scott Reed, Dragomir
  Anguelov, Dumitru Erhan, Vincent Vanhoucke, and Andrew Rabinovich.
\newblock Going deeper with convolutions.
\newblock In {\em CVPR}, pages 1--9, 2015.

\bibitem{szegedy2014intriguing}
Christian Szegedy, Wojciech Zaremba, Ilya Sutskever, Joan Bruna, Dumitru Erhan,
  Ian Goodfellow, and Rob Fergus.
\newblock Intriguing properties of neural networks.
\newblock In {\em ICLR}, 2014.

\bibitem{tan2020efficientdet}
Mingxing Tan, Ruoming Pang, and Quoc~V Le.
\newblock Efficientdet: Scalable and efficient object detection.
\newblock In {\em CVPR}, pages 10781--10790, 2020.
\newblock {(Apache-2.0)}:
  \url{https://github.com/google/automl/tree/master/efficientdet}.

\bibitem{torralba2011unbiased}
Antonio Torralba and Alexei~A Efros.
\newblock Unbiased look at dataset bias.
\newblock In {\em CVPR}, pages 1521--1528, 2011.

\bibitem{vaicenavicius19evalcal}
Juozas Vaicenavicius, David Widmann, Carl Andersson, Fredrik Lindsten, Jacob
  Roll, and Thomas Sch\"{o}n.
\newblock Evaluating model calibration in classification.
\newblock In {\em Artificial Intelligence and Statistics}, volume~89, pages
  3459--3467, 2019.

\bibitem{wang2020cspnet}
Chien-Yao Wang, Hong-Yuan Mark~Liao, Yueh-Hua Wu, Ping-Yang Chen, Jun-Wei
  Hsieh, and I-Hau Yeh.
\newblock Cspnet: A new backbone that can enhance learning capability of cnn.
\newblock In {\em CVPRW}, pages 1571--1580, 2020.

\bibitem{Wang_2020visualbias}
Zeyu Wang, Klint Qinami, Ioannis~Christos Karakozis, Kyle Genova, Prem Nair,
  Kenji Hata, and Olga Russakovsky.
\newblock Towards fairness in visual recognition: Effective strategies for bias
  mitigation.
\newblock In {\em CVPR}, pages 8919--8928, 2020.

\bibitem{winkler1968good}
Robert~L Winkler and Allan~H Murphy.
\newblock “good” probability assessors.
\newblock {\em Journal of Applied Meteorology and Climatology}, 7(5):751--758,
  1968.

\bibitem{xie2017resnext}
Saining Xie, Ross Girshick, Piotr Doll{\'a}r, Zhuowen Tu, and Kaiming He.
\newblock Aggregated residual transformations for deep neural networks.
\newblock In {\em CVPR}, pages 1492--1500, 2017.

\bibitem{zadrozny200histbin}
Bianca Zadrozny and Charles Elkan.
\newblock Obtaining calibrated probability estimates from decision trees and
  naive bayesian classifiers.
\newblock In {\em ICML}, volume~1, pages 609--616, 2001.

\bibitem{zadrozny2002transforming}
Bianca Zadrozny and Charles Elkan.
\newblock Transforming classifier scores into accurate multiclass probability
  estimates.
\newblock In {\em ACM SIGKDD international conference on Knowledge discovery
  and data mining}, pages 694--699, 2002.

\bibitem{zemel2013learningfair}
Rich Zemel, Yu Wu, Kevin Swersky, Toni Pitassi, and Cynthia Dwork.
\newblock Learning fair representations.
\newblock In {\em ICML}, pages 325--333, 2013.

\bibitem{zhao2017men}
Jieyu Zhao, Tianlu Wang, Mark Yatskar, Vicente Ordonez, and Kai-Wei Chang.
\newblock Men also like shopping: Reducing gender bias amplification using
  corpus-level constraints.
\newblock In {\em Conference on Empirical Methods in Natural Language
  Processing (EMNLP)}, 2017.

\bibitem{zhou2021cnet2}
Xingyi Zhou, Vladlen Koltun, and Philipp Kr{\"a}henb{\"u}hl.
\newblock Probabilistic two-stage detection.
\newblock {\em arXiv preprint arXiv:2103.07461}, 2021.
\newblock {(Apache-2.0)}: \url{https://github.com/xingyizhou/CenterNet2}.

\bibitem{zhou2019objects}
Xingyi Zhou, Dequan Wang, and Philipp Kr{\"a}henb{\"u}hl.
\newblock Objects as points.
\newblock {\em arXiv preprint arXiv:1904.07850}, 2019.
\newblock {(MIT License)}: \url{https://github.com/xingyizhou/CenterNet}.

\bibitem{zhu2019deformable}
Xizhou Zhu, Han Hu, Stephen Lin, and Jifeng Dai.
\newblock Deformable convnets v2: More deformable, better results.
\newblock In {\em CVPR}, pages 9308--9316, 2019.

\end{thebibliography}
}
\clearpage
\appendix

\section{Maximizing \ap - Extended} \label{sec:a_max_ap}
Taking an extended look at the formal proof we start again with the expected \ap: 
\begin{equation*}
\E_{\tp{}} [\ap] = \E_{\tp{}} \Bigg[ \sum_{i=1}^{N} \precision(i)  \cdot \Delta \recall(i) \Bigg].
\label{eq:a_exp_ap0}
\end{equation*}
Substituting precision and recall and our stochastic indicator variable \tp{} we first get:
\begin{equation*}
\E_{\tp{}} [\ap] = \E_{\tp{}} \Bigg[ \sum_{i=1}^{N} \bigg( \frac{\sum_{k=1}^{i}(\tp{k})}{i} \cdot \frac{\tp{i}}{|\gtdetects|} \bigg) \Bigg].
    \label{eq:a_exp_ap1}
\end{equation*}
We can move \tp{i} out of the inner sum:
\begin{equation*}
\E_{\tp{}} [\ap] = \E_{\tp{}} \Bigg[ \sum_{i=1}^{N} \bigg( \frac{\sum_{k=1}^{i-1}(\tp{k}) + \tp{i}}{i} \cdot \frac{\tp{i}}{|\gtdetects|} \bigg) \Bigg].
    \label{eq:a_exp_ap1_1}
\end{equation*}
We assume independence of \tp{n} and \tp{m} for every $m,n$ with $m \neq n$. This is actually only the case if the detections \detect{n} and \detect{m} don't try to detect the same ground truth object. The introduced error, however, is minuscule on a large dataset the number of detections for the same ground truth object are significantly smaller than the overall number of detections. The number of detections for the same object are further decreased through \gls{nms}.\\  The number of detections $|\gtdetects|$ is constant and can be moved to the front. Since $\tp{i} \sim \operatorname{Bernoulli}(\probtp{i})$ it follows that $(\tp{i})^2 = \tp{i}$.
\begin{equation*}
\E_{\tp{}} [\ap] = \frac{1}{|\gtdetects|}  \sum_{i=1}^{N} \bigg(\frac{\sum_{k=1}^{i-1}(\probtp{k}) + 1}{i} \cdot \probtp{i}\bigg).
    \label{eq:a_exp_ap2}
\end{equation*}
First we move the inner sum to the back,
\begin{equation*}
\E_{\tp{}} [\ap] = \frac{1}{|\gtdetects|}  \sum_{i=1}^{N} \bigg(\frac{\probtp{i}}{i} +\probtp{i}\frac{\sum_{k=1}^{i-1}(\probtp{k})}{i} \bigg)
    \label{eq:a_exp_ap2_1}
\end{equation*}
and then reformulate it. First we split it into the first sum and double sum:
\begin{equation*}
\E_{\tp{}} [\ap] = \frac{1}{|\gtdetects|}  \sum_{i=1}^{N} \bigg(\frac{\probtp{i}}{i} \bigg) + \sum_{i=1}^{N} \sum_{k=1}^{i-1}\frac{(\probtp{i}\cdot\probtp{k})}{i}. 
    \label{eq:a_exp_ap2_2}
\end{equation*}
Then we can switch the sums and the limits:
\begin{equation*}
\E_{\tp{}} [\ap] = \frac{1}{|\gtdetects|}  \sum_{i=1}^{N} \bigg(\frac{\probtp{i}}{i} \bigg) + \sum_{i=1}^{N} \sum_{k=i+1}^{N}\frac{(\probtp{i}\cdot\probtp{k})}{k} .
    \label{eq:a_exp_ap2_3}
\end{equation*}
Then we re-combine the first sum with the outer of the second sums:
\begin{equation*}
\E_{\tp{}} [\ap] =\frac{1}{|\gtdetects|} \sum_{i=1}^{N} \underbrace{\bigg( \frac{\probtp{i}}{i} + \probtp{i} \sum_{k=i+1}^{N}\frac{\probtp{k}}{k} \bigg)}_{h_i(\probtp{i}, \probtp{})}.
    \label{eq:a_exp_ap3}
\end{equation*}
Here we see that $h_i(l,\probtp{}) > h_{i+1}(l,\probtp{})$ for $i \in \mathbb{N}$ and $l \in (0,1]$. This can be seen more clearly if we split $h_i(l,\probtp{}) $ into the two components of its sum $(\text{I})$ and$(\text{II})$, 
\begin{equation*}
    h_i(\probtp{i}, \probtp{}) = \underbrace{\vphantom{\sum_{k=i+1}^{N}}\frac{\probtp{i}}{i}}_{(\text{I})} + \underbrace{\probtp{i} \sum_{k=i+1}^{N}\frac{\probtp{k}}{k}}_{(\text{II})},
\end{equation*}
for the $(\text{I})$ it is obvious that for any $l \in (0,1]$ and $i \in \mathbb{N}$:
\begin{equation*}
 \frac{l}{i} > \frac{l}{i+1}.
    \label{eq:a_exp_ap3_1}
\end{equation*}
For the second term $(\text{II})$ we can see that for any $i$ it can be split as follows:
\begin{equation*}
  l \sum_{k=i+1}^{N}\frac{\probtp{k}}{k} = \underbrace{l \frac{\probtp{i+1}}{i+1}}_{(\text{III})} + l \sum_{k=i+2}^{N}\frac{\probtp{k}}{k},
    \label{eq:a_exp_ap3_2}
\end{equation*}
where $(\text{III})$ is $>0$ for $l \in (0,1]$ and $i \in \mathbb{N}$ and the second term is $(\text{II})$ of $h_{i+1}(l,\probtp{})$. Which proves that $h_i(l,\probtp{})$ is strictly larger than $h_{i+1}(l,\probtp{})$ in the relevant intervals $l \in (0,1]$ and $i \in \mathbb{N}$.
So the expected \ap is a sum of functions $h$, that given the same input value have strictly decreasing output for larger values of $i$. It can thereby be maximized for some fixed set of \detects, by sorting the detections by their \probtp{}. Since the detections are already sorted according to their \confidence{} for the valuation we need to ensure that this also sorts \probtp{} i.e. that the confidence calibration curve is monotonic:
\begin{equation*}
	\probtp{n} < \probtp{m} \; \forall\ n,m\, | \,  \confidence{n} < \confidence{m}.
    \label{eq:a_ap_maximized}
\end{equation*}
\section{More Optimization Metrics} \label{sec:a_more_optim_mets}
\begin{figure}[ht!] 
       \includegraphics[width={\columnwidth}]{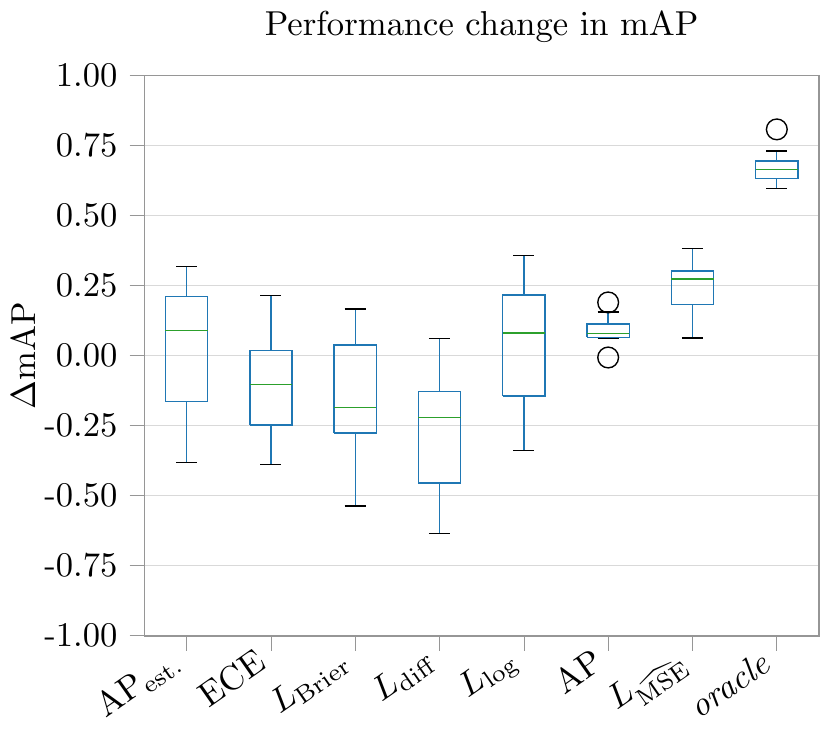}
       \caption{\textbf{Performance change in \map for extended optimization metrics.} CenterNet calibrated on 60\% of \gls{coco} validation split, evaluated on the remaining 40\% with 10 random splits. The box ranges from the lower to upper quantile values, the green line is the median performance change.} 
    \label{fig:optimmets_map}
\end{figure}

\begin{figure}[ht!] 
       \includegraphics[width={\columnwidth}]{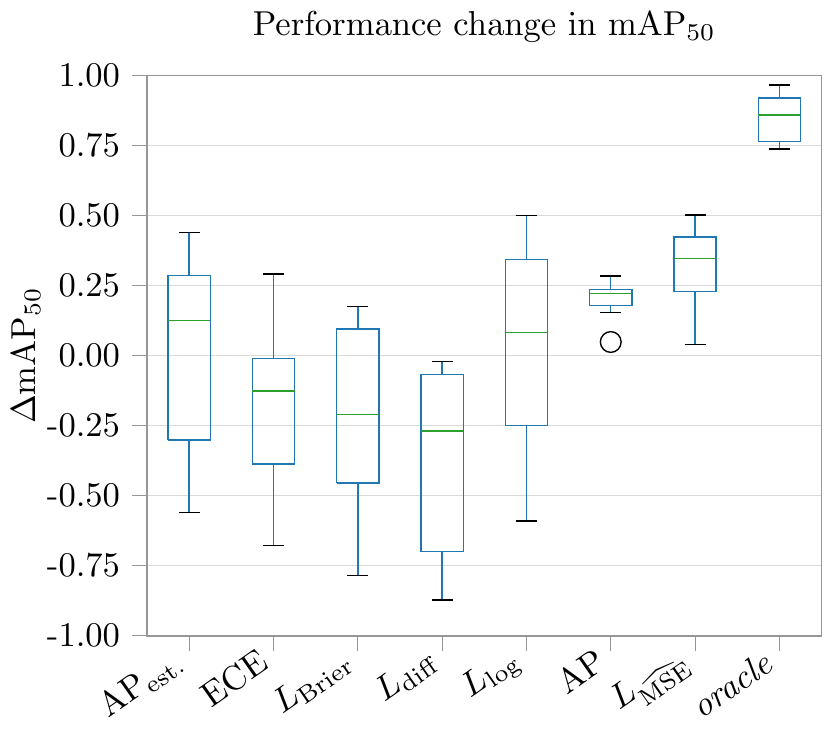}
       \caption{\textbf{Performance change in \mapfifty for extended optimization metrics.} Same settings as in \cref{fig:optimmets_map} }   
    \label{fig:optimmets_mapfif}
\end{figure}

There are a range of metrics that could be explored for the optimization of the bin size parameter space. The explored \ap, \lossbrier, \losslog, and \lossmse each have a good theoretical justification for usage in this application. We explore some of the possible metrics which we did not include in the main section, and give justification for their exclusion. \\
\shortsec{Absolute Difference.}
The absolute difference, or absolute deviations, could be considered a reasonable choice besides \lossbrier and \losslog. It is calculated as
\begin{equation*}
	\lossdiff = \frac{1}{N} \sum_{i=1}^{N} | \confidence{i} - \tpindic{i} |, 
    \label{eq:_diff}
\end{equation*}
but, in contrast to \lossbrier and \losslog, it is not a proper scoring rule \cite{gneiting2007strictly}. It is not minimized for $\confidence{i} = \probtp{i}$, but rather by the majority label, \ie by $\confidence{i} = 1$ for $\probtp{i} > 0.5$ and $\confidence{i} = 0$ for $\probtp{i} < 0.5$. Unsurprisingly, it performs even worse than the proper scoring rules for the performance measured in \map (\cf \cref{fig:optimmets_map}) and \mapfifty (\cf \cref{fig:optimmets_mapfif}) \\
\shortsec{Expected Calibration Error.} Since our goal is to perform a conditional confidence calibration a intuitive choice for the optimization metric is the \gls{ece} \cite{naeini2015ece_bayesbin}. If we let $\calibfunc_{1,\confbin{}}$ be the un-modified histogram binning with \confbin{} confidence bins, the \gls{ece} is calculated as:
\begin{equation*}
	\text{ECE} = \frac{1}{N} \sum_{i=1}^{N} | \confidence{i} - \calibfunc_{1,\confbin{}}(\detect{i})|.
    \label{eq:a_ece}
\end{equation*}
The \gls{ece} is also a proper scoring rule \cite{gneiting2007strictly}, but it also has its limitations in general~\cite{nixon2019measuring_ace} and for this application: It only tries to captures the calibration error not the conditional calibration bias. For the parameter optimization we follow~\cite{guo2017on_calib_nn} and set the number of confidence bins to $\confbin{}=10$. We calculate the \gls{ece} separately for each class because we want search for the class-wise optimal parameters. Because of the described drawbacks the \gls{ece} does not perform well as a optimization metric (\cf \cref{fig:optimmets_map}).
\\
\shortsec{Estimated \ap.} Instead of the 11-point maximum interpolated \ap metric used by the \gls{coco} benchmark-evaluation we could use the the \ap computed over all the available detections for each class:  
\begin{equation*}
	\ap = \sum_{i=1}^{N} \precision(i)  \cdot \Delta \recall(i).
\label{eq:a_AP_calculated}
\end{equation*}
To distinguish it from the \gls{coco}-benchmark \ap metric we refer to it as \apest. On average, \apest performs almost as good as \ap when used as the optimization metric. It is, however, also more susceptible to outliers and can thereby sometimes severely degrade the performance on the hold-out set (\cf \cref{fig:optimmets_map}). 
\section{Effect of Model Size} \label{sec:a_model_size}
We observed that the effectiveness of approach appeared to be negatively correlated with the object detectors model size and performance. We verify the observation on the EfficientDet \cite{tan2020efficientdet}, which is available in 8 different size and performance versions. This ensures that there are no other influences, such as loss functions, augmentations or similar factors. The trend largely holds for the different EfficientDet model variants, but there ares some minor outliers (\cf \cref{tab:modelsize}).
\begin{table*}[ht!]
    \centering
    \newcolumntype{s}{>{\columncolor[HTML]{EDEDED}} l}
\begin{tabular}{cclsls}
\toprule
\multicolumn{2}{c}{Search Space}     & \multicolumn{2}{c}{Calibration (train)} & \multicolumn{2}{c}{\textit{Oracle}}   \\\cmidrule(l){1-2}\cmidrule(l){3-4} \cmidrule(l){5-6}
Box Bins &   Confidence Bins                                   &   \map   &   \mapfifty  &   \map   &   \mapfifty     \\ 
\midrule\midrule

\boxbin{0} &\confbin{0} &                                       40.52\change{+0.23} & 59.39\change{+0.30}  & 40.78\change{+0.49} & 59.78\change{+0.69}  \\
\boxbin{0} &$\confbin{1}$ &                                     40.43\change{+0.14} & 59.48\change{+0.39}  & 40.88\change{+0.59} & 59.88\change{+0.79}  \\
$\boxbin{0} \cup \boxbin{1}$ & \confbin{0}          &           40.61\change{+0.32} & 59.57\change{+0.48}  & 40.89\change{+0.60} & 59.89\change{+0.80} \\
$\boxbin{0} \cup \boxbin{1}$ &$\confbin{0} \cup \confbin{1}$ &  40.61\change{+0.32} & 59.58\change{+0.49}  & 40.99\change{+0.70} & 59.99\change{+0.90} \\
\midrule
\multicolumn{2}{c}{Baseline} &                                  40.29               & 59.09                & 40.29               & 59.09  \\
\bottomrule
\end{tabular}
    \caption{\textbf{Influence of parameter search spaces on performance gain.} Performance of calibrated CenterNet detector with parameters optimized with \lossmse metric and \textit{oracle} evaluation. Calibration on \gls{coco} train split, evaluation on validation data. Larger search space enables larger performance gains, but excluding smaller sized confidence bins from the search space (\confbin{0}) can reduce \map when optimizing for the \lossmse metric.}
    \label{tab:a_searchspace}
\end{table*}
\begin{table*}[ht!]
    \centering
    \begin{tabular}{cccllll}
\toprule
Version      & \#Parameters & \#FLOPs 
& Calibration &\map & \mapfifty \\
\midrule
\multirow{3}{*}{EfficientDet-D0} & \multirow{3}{*}{3.9M}  & \multirow{3}{*}{2.54B}&
 -  & 34.24                 &  52.48 \\   
                  &     & & train    & 34.30\change{+0.06}   &  52.62\change{+0.14} \\  
                 &   &    & \textit{oracle} & 34.50\change{+0.26}                 &  53.00\change{+0.52}\\
\midrule
\multirow{3}{*}{EfficientDet-D1}& \multirow{3}{*}{6.6M}  & \multirow{3}{*}{6.10B} &
 -  & 40.09                 &  58.85 \\  
                 &      & &  train   & 40.16\change{+0.07}   &  58.95\change{+0.10} \\ 
                  &  &    & \textit{oracle} & 40.33\change{+0.24}                 &  59.30\change{+0.45}\\
\midrule
\multirow{3}{*}{EfficientDet-D2}& \multirow{3}{*}{8.1M}  & \multirow{3}{*}{11.0B} &
 -  & 43.38                 &  62.52 \\                
                &       & &  train   & 43.42\change{+0.04}   &  62.64\change{+0.12} \\  
                 &   &    & \textit{oracle} & 43.61\change{+0.23}                 &  62.99\change{+0.47}\\
\midrule
\multirow{3}{*}{EfficientDet-D3}& \multirow{3}{*}{12.0M}  & \multirow{3}{*}{24.9B}&
 -  & 47.05                 &  65.86 \\ 
               &    & &  train   & 47.08\change{+0.03}   &  65.90\change{+0.04} \\ 
              &  &    & \textit{oracle} & 47.23\change{+0.18}                 &  66.18\change{+0.32}\\
\midrule
\multirow{3}{*}{EfficientDet-D4}& \multirow{3}{*}{20.7M} & \multirow{3}{*}{55.2B} &
 -  & 49.15                 &  68.24 \\      
                &       & &  train   & 49.16\change{+0.01}   &  68.27\change{+0.03} \\  
                &    &    & \textit{oracle} & 49.33\change{+0.18}                 &  68.58\change{+0.34}\\
\midrule
\multirow{3}{*}{EfficientDet-D5}& \multirow{3}{*}{33.7M} & \multirow{3}{*}{130B}  &
 -  & 51.03                 &  70.09 \\                
                &         & &  train & 51.08\change{+0.05}   &  70.16\change{+0.07} \\  
                &         & & \textit{oracle} & 51.25\change{+0.22}                 &  70.45\change{+0.36}\\
\midrule
\multirow{3}{*}{EfficientDet-D6}& \multirow{3}{*}{51.9M} & \multirow{3}{*}{226B}  &
 -  & 51.99                 &  70.94 \\                
                &         & &  train & 52.00\change{+0.01}   &  70.98\change{+0.04} \\  
                &         & & \textit{oracle} & 52.17\change{+0.18}                 &  71.27\change{+0.33}\\
\midrule
\multirow{3}{*}{EfficientDet-D7}& \multirow{3}{*}{51.9M} & \multirow{3}{*}{325B}  &
 -  & 53.06                 &  72.12 \\                
                &         & &  train & 53.05\change{-0.01}   &  72.14\change{+0.02} \\  
                &         & & \textit{oracle} & 53.21\change{+0.15}                 &  72.42\change{+0.30}\\
\bottomrule
\end{tabular}

    \caption{\textbf{Influence of calibration method on different sized versions of EfficientDet \cite{tan2020efficientdet}.} Ordered by increasing model size: Calibration and parameter optimization on \gls{coco} train, evaluated on validation data. The calibration is not very effective, its impact decreases with increasing model size.}
    \label{tab:modelsize}
\end{table*}
\section{Parameter Search Space} \label{sec:a_search_space}
We chose a fixed search space of $\boxbin{0} = \{2, 3, 4, 5, 6\}$ and $\confbin{0} = \{4, 5, 6, 8, 10, 12, 14\}$, which we kept constant to have comparable results for all detectors and methods. We now take a closer look at the influence the search space has on the performance. We define two additional sets $\boxbin{1} = \{8, 10, 12, 14, 20\}$ and $\confbin{1} = \{14, 16, 18, 20, 24, 28, 34, 40, 50\}$ and explore different combinations of the four sets for the parameter search space. The results show that of course a larger search space increases the performance gains (\cf \cref{tab:a_searchspace}). If, however, the search space excludes low values for the number of confidence bins like in \confbin{1}, the performance for categories with few detections can be decreased.

\end{document}